\documentclass{article}
\usepackage{spconf,amsmath,graphicx}
\usepackage{balance}
\usepackage{epsfig,url}
\usepackage{color}
\usepackage{multirow}
\usepackage{booktabs}
\usepackage{array}

\newcommand\etal{{\it et~al.~}}
\newcommand\ie{{\it i.e.}}

\newcommand\eg{{\it e.g.}}

\title{Adaptive Scenario Discovery for Crowd Counting}
\name{Xingjiao Wu$^{1}$, Yingbin Zheng$^{2,3}$, Hao Ye$^{2,3}$, Wenxin Hu$^{1*}$\thanks{$^*$Corresponding author.}, Jing Yang$^{1}$, Liang He$^{1}$}
\address{$^1$East China Normal University, Shanghai, China\\
$^2$Shanghai Advanced Research Institute, CAS~~$^3$Videt Tech.\\
{\small 52184506007@stu.ecnu.edu.cn, \{yingbin.zheng, hao.ye\}@videt.cn, wxhu@cc.ecnu.edu.cn, \{jyang, lhe\}@cs.ecnu.edu.cn}}

\begin{document}

\maketitle

\begin{abstract}

Crowd counting, \ie, estimation number of the pedestrian in crowd images, is emerging as an important research problem with the public security applications.
A key component for the crowd counting systems is the construction of counting models which are robust to various scenarios under facts such as camera perspective and physical barriers.
In this paper, we present an adaptive scenario discovery framework for crowd counting. The system is structured with two parallel pathways that are trained with different sizes of the receptive field to represent different scales and crowd densities. After ensuring that these components are present in the proper geometric configuration, a third branch is designed to adaptively recalibrate the pathway-wise responses by discovering and modeling the dynamic scenarios implicitly. Our system is able to represent highly variable crowd images and achieves state-of-the-art results in two challenging benchmarks.

\end{abstract}


\section{Introduction}
\label{sec:intro}

Counting is the process of estimating the number of a particular object.
With the expansion of urban population and the convenience of modern transportation, it is common to have large crowds in specific events or scenarios, and crowd counting from images or videos becomes crucial for applications ranging from traffic control to public safety.

\begin{figure}[t]
  \centering
  \includegraphics[width=.98\linewidth]{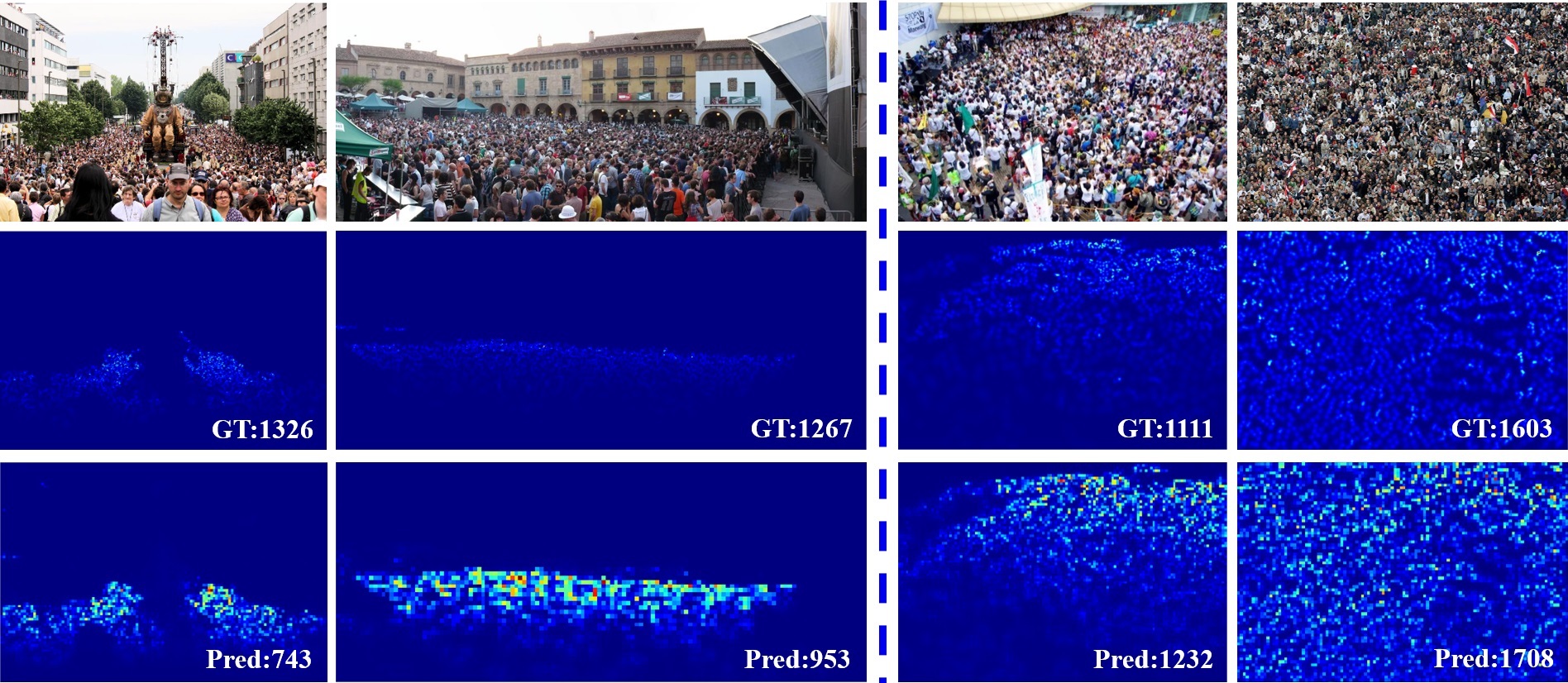}
  \caption{The crowd images from the ShanghaiTech dataset~\cite{Zhang_2016_CVPR} and their crowd counting prediction by CSRNet~\cite{Li2018CSRNet}.}
  \label{fig:motivation}
  \vspace{-0.1in}
\end{figure}

Previous methods of crowd counting may be roughly divided into two categories: detection-based and regression-based.
Detection-based methods have been studied with the pedestrian detectors \cite{2011CVPR_MWang,Stewart2016EndtoEndPD}.
However, it is challenging for these methods to model a very dense crowd or crowd in a clustered environment. The regression-based approaches are firstly proposed in \cite{Idrees2013Multi}. With the recent development of the convolutional neural network (CNN), the regression framework by estimation of the density maps has been widely used. Compared with the system employing a single CNN regressor (\eg, \cite{onoro2016}), the networks with multiple columns/branches learn more contextual information and achieve excellent performance~\cite{Zhang_2016_CVPR, Sam_2017_CVPR, Sindagi2017Generating, Li2018CSRNet}.
Although different receptive fields are usually applied in multiple branches, it is difficult to represent highly variable crowd images.
%As shown in Fig. \ref{fig:motivation}, there still exist gaps between the ground-truth and prediction for some crowd images.
There still exist gaps between the ground-truth and prediction for some crowd images (some examples are shown in Fig. \ref{fig:motivation}).
We also observe that the images under similar scenario seem to have the same prediction pattern: the images with the lower camera viewpoints and more backgrounds usually achieve smaller counting prediction than the ground-truth (Fig. \ref{fig:motivation}-Left), while these with high viewpoint get larger predicted values (Fig. \ref{fig:motivation}-Right).

The central issue addressed in this paper is the following: \emph{Can we design a model to discover the scenarios and modeling the crowd images simultaneously?} One intuitive idea is to add the number of network branch with well-designed convolution filters. The limitations are, the CNN model will be difficult to train with the current crowd counting datasets, and it is also hard to directly define the scenarios. In
this paper, we present an adaptive scenario discovery framework for crowd counting. Our network adopts the VGG model~\cite{Simonyan15} as the backbone and is structured with two parallel pathways that are trained with different sizes of the receptive field to serve different scales and crowd densities. We consider the scenario as a linear combinational of two pathway with the discretized weights and design a third adaption branch to learn this scenario aware responses and discover the scenarios implicitly.

\newpage
Our contributions are summarized as follows.
\begin{itemize}
\item From the perspective of scenario discovery, a novel adaptive framework for crowd counting is proposed. Different from previous multiple columns/branches frameworks, ours has the ability to represent highly variable crowd images with two branches by incorporating the discretized pathway-wise responses.
\item We apply our framework to the ShanghaiTech~\cite{Zhang_2016_CVPR} and UCF\_CC\_50~\cite{Idrees2013Multi} crowd counting datasets, and find that it outperforms the state-of-the-art approaches.
\end{itemize}

\subsection{Related Work}
\label{sec:related}

Numerous efforts have been devoted to the design of crowd counting models. Detail survey of the recent progress can be found in \cite{sindagi2018survey}. In this section, we mainly discuss literature on the models with multiple branches representation, which are more related to this work.
In \cite{Zhang_2016_CVPR}, Zhang \etal proposed the MCNN by using three columns of convolutional neural networks with filters of different sizes.
Sam \etal \cite{Sam_2017_CVPR} proposed the Switching-CNN, which decoupled the three columns into separate CNN (each trained with a subset of the patches), and a density selector is designed to utilize the structural and functional differences.
Several works have studied the context information of the crowd images under multiple branch setting. For instance, Sindagi \etal~\cite{Sindagi2017Generating} applied local and global context coding to population count density estimation, and
Zhang \etal\cite{Zhang2018Crowd} proposed a scale-adaptive CNN architecture with a backbone of fixed small receptive fields. Another work related to ours is the CSRNet~\cite{Li2018CSRNet}, where convolutional neural networks with dilation operations were employed after the backbone of the pre-trained deep model.

These existing approaches construct density estimation models with multiple branches to represent different receptive fields or scales. Our framework also follows the general process, with the design of one branch representing the dense prediction and another for the relative sparse crowds. However, instead of using the fix branch weights or selecting one explicitly column, we adopt the learning of branch weights. Responses of the dense and sparse pathways are adaptively recalibrated by a third branch, which explicitly models interdependencies between pathways. Moreover, with the discretization of these pathway-wise responses, the crowd scenarios are implicitly discovered and respond to different crowd images in a highly scenario-specific manner. The whole framework can be end-to-end trained, and as will be shown in the experiments, it is more accurate compared to previous approaches.

\section{Framework}
\label{sec:related}

\begin{figure}[t]
  \centering
  \includegraphics[width=\linewidth]{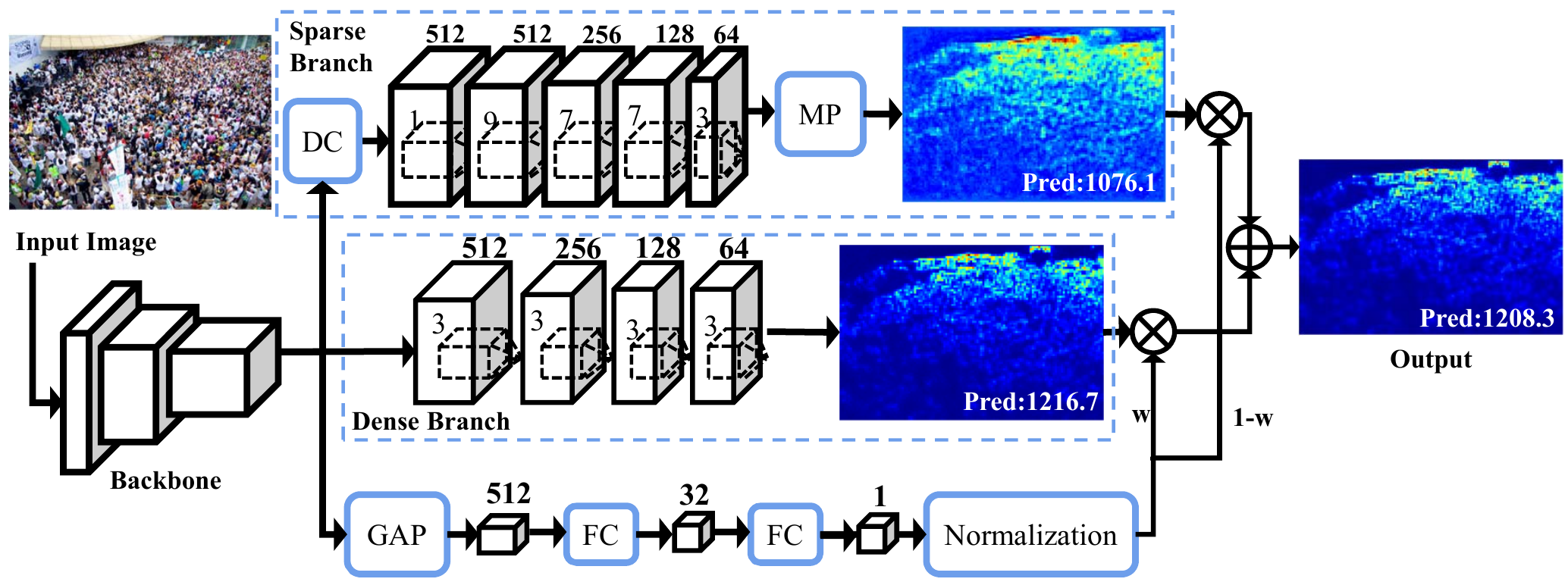}
  %\vspace{-0.1in}
  \caption{Network structure of proposed framework. DC, MP, GAP, and FC indicate deconvolution layer, max pooling, global average pooling, and full-connected layer respectively.}
  \label{fig:framework}
  %\vspace{-0.1in}
\end{figure}

The overall architecture of our framework is illustrated in Fig. \ref{fig:framework}. We start by introducing the design of adaptive scenario discovery, followed by implementation details of the framework.

\subsection{Adaptive Scenario Discovery}

The selection of a suitable network structure is important to the success of a crowd counting system. There are generally two categories of networks: either it is with a new design of the structure and learned from scratch (\eg, \cite{Zhang_2016_CVPR,Shen_2018_CVPR}), or the model is transferred from part of a pre-trained network (\eg, \cite{Li2018CSRNet,Shi_2018_CVPR}). In this paper, our framework belongs to the second case, by employing the convolutional layers of a VGG-16 model \cite{Simonyan15} pre-trained from the ImageNet dataset \cite{deng2009imagenet} and fine-tuned with the crowd images. We choose this strategy for the outstanding performance of the model in crowd counting as well as other computer vision tasks, and the results in the evaluation also confirm the effects of the pre-trained model.

Our counting network consists of two parallel pathways after the backbone module.
The first pathway starts with a deconvolution layer that amplifies the inputs, and then a few convolutional layers with larger receptive fields are used, followed by a $2\times2$ max pooling. This pathway is designed to model the high congested scenario with \emph{dense} crowd, and the second pathway is for the \emph{sparse} scenario. The convolution filers in this subnet are with a size of $3\times3$. Note that the concept of dense or sparse is relative and both pathways can output a density map.

There are several approaches to fuse the density maps, and here we would like to use a dynamic weighting strategy. Inspired by the excitation operation in SENet~\cite{Hu2017SqueezeandExcitationN}, we propose the \emph{adaption} branch. The outputs of the last convolutional layer in the backbone go through a global average pooling and two fully-connected layers and then have an initial response $w$. We expect $w$ to adaptively recalibrate the weight of the dense and sparse pathways, therefore we normalize it into the interval of [0,1) with the following formula:
\begin{equation}
w^* = \arctan({\rm sigmod}(w)) \times \frac{2}{\pi}
\label{equ:atan}
\end{equation}
Experiments on Section 3.2 will show the effect compared with the single branch or average fusion. However, we find that the convergence speed of this architecture is slow, probably due to the small size of the crowd counting dataset but the continuous response.

Our solution is to divide the response value into bins, by borrowing the idea from traditional visual features such as color histogram~\cite{stricker1995similarity}, SIFT~\cite{lowe2004distinctive}, and HoG~\cite{dalal2005histograms}. The benefits of discretization are two-folder. First, the model itself is easier to train and converge. Second, similar attributes are significantly observed from the images within the same bin (see Fig. \ref{fig:scenario}), indicating that discretization operation is able to implicitly discovering the dynamic scenarios.

\subsection{Implementation Details}
\textbf{Ground Truth Generation.}
We follow \cite{Li2018CSRNet} to generate the density maps from ground truth.
the density map $F(x)$ is generated with the formula:
\begin{equation}
F(x) = \sum\limits_{i = 1}^N {\delta (x - {x_i})}  \times {G_{\sigma_i}}(x)
\label{equ:F}
\end{equation}
where $x_i$ is a targeted object in the ground truth $\delta$ and $G_{\sigma_i}(\cdot)$ is a Gaussian kernel with standard deviation of $\sigma_i$. For the datasets with high congested scene (such as ShanghaiTech Part A~\cite{Zhang_2016_CVPR} and UCF\_CC\_50~\cite{Idrees2013Multi}), $F(x)$ is defined as a geometry-adaptive kernel with $\sigma_i= \beta {\bar d_i}$. Here ${\bar d_i}$ is the average distance of $k$ nearest neighbors of targeted object $x_i$. For low congested scene (\ie, ShanghaiTech Part B~\cite{Zhang_2016_CVPR}), we set $\sigma_i=15$.

\vspace{0.08in}
\noindent\textbf{Training Details.}
We define the loss function as follows:
\begin{equation}
\begin{array}{l}
L(\Theta ) = \frac{1}{2N}\sum\limits_{i = 1}^N {\left\| {\mathcal{F}({X_i};\Theta ) - {F(X_i)}} \right\|_2^2}
\end{array}
\label{equ:loss}
\end{equation}
where $F(X_i)$ is the ground truth density map of image $X_i$ from Equ. (\ref{equ:F}) and $\mathcal{F}({X_i};\Theta)$ is the estimated density map of $X_i$ with the parameters $\Theta$ learned by the proposed network.

To ensure the spatial feature and the context of the crowd images, we do not extract the image patches for data augmentation. And there is also no additional image copy/conversion enhancement.
During training, we employ the stochastic gradient descent (SGD) for its good generalization ability.

\section{Evaluations}
\label{sec:exp}

\begin{table}[t]
\centering
\caption{Comparison with the state-of-the-arts on the benchmarks. Part A and Part B indicate ShanghaiTech Part A and Part B, respectively.}
\label{table:mae}
\small{
\begin{tabular}{p{.95in}p{0.22in}p{0.28in}p{0.22in}p{0.22in}p{0.22in}p{0.22in}}
\hline
\multicolumn{1}{c}{\multirow{2}{*}{Method}} & \multicolumn{2}{c}{Part A} & \multicolumn{2}{c}{Part B} & \multicolumn{2}{c}{UCF\_CC\_50} \\  \cline{2-7}
\multicolumn{1}{c}{} & MAE & MSE         & MAE          & MSE      & MAE          & MSE \\  \cline{1-7}
Zhang et al.~\cite{Zhang2015Cross}  & 181.8        & 277.7        & 32.0           & 49.8    & 467.0   & 498.5 \\
MCNN~\cite{Zhang_2016_CVPR} & 110.2        & 173.2        & 26.4         & 41.3   & 377.6 & 509.1 \\
\footnotesize{Cascaded-MTL}~\cite{Sindagi2017CNN} & 101.3        & 152.4        & 20.0         & 31.1  & 322.8 & 397.9\\
\footnotesize{Switching-CNN}~\cite{Sam_2017_CVPR} & 90.4         & 135.0          & 21.6         & 33.4   & 318.1 & 439.2\\
DAN~\cite{wang2018crowd}&88.5&147.6& 17.6& 26.8&234.5&289.6\\
CP-CNN~\cite{Sindagi2017Generating} & 73.6         & 106.4        & 20.1         & 30.1 & 295.8 & 320.9\\
Huang \etal~\cite{Huang2017Body} & -       & -        & 20.2         & 35.6 & 409.5 & 563.7 \\
D-ConvNet~\cite{Shi_2018_CVPR} & 73.5         & 112.3          & 18.7         & 26.0 & 288.4 & 404.7\\
ACSCP~\cite{Shen_2018_CVPR}                   & 75.7         & 102.7          & 17.2        & 27.4 & 291.0 & 404.6 \\
DecideNet~\cite{Liu2018DecideNet}&-&-& 20.8& 29.4&-&-\\
SaCNN~\cite{Zhang2018Crowd} & 86.8         & 139.2         & 16.2         & 25.8 & 314.9 & 424.8 \\
CSRNet~\cite{Li2018CSRNet} & 68.2         &  115.0         & 10.6         &  16.0 & 266.1 & 397.5  \\
ASD [ours]  &\textbf{65.6} &\multicolumn{1}{r}{\textbf{98.0}}& \multicolumn{1}{r}{\textbf{8.5}}  & \textbf{13.7} &\textbf{196.2}  &\textbf{270.9}  \\
\hline
\end{tabular}
}
\end{table}

We conduct the experiments on the ShanghaiTech dataset~\cite{Zhang_2016_CVPR} and the UCF\_CC\_50 dataset~\cite{Idrees2013Multi}. The ShanghaiTech dataset~\cite{Zhang_2016_CVPR} is divided into Part A and Part B. ShanghaiTech  Part A contains 482 crowd images with 300 training images and 182 testing images, and the average number of the pedestrian is 501. ShanghaiTech Part B is with 716 images (400 training and 316 testing). The resolution of the images are fixed with $768\times1024$ pixels, and the pedestrian number is generally smaller than Part A with an average number of 123. The UCF\_CC\_50 dataset~\cite{Idrees2013Multi} contains 50 images with high crowd density. The images vary in the number of pedestrians, with a range of 94 to 4,543. For both datasets, we follow the standard experimental protocols, and mean absolute error (MAE) and mean squared error (MSE) is reported as the evaluation metric. We implement our framework based on PyTorch~\cite{pytorch}.

\begin{figure}[t]
  \centering
  \includegraphics[width=.99\linewidth]{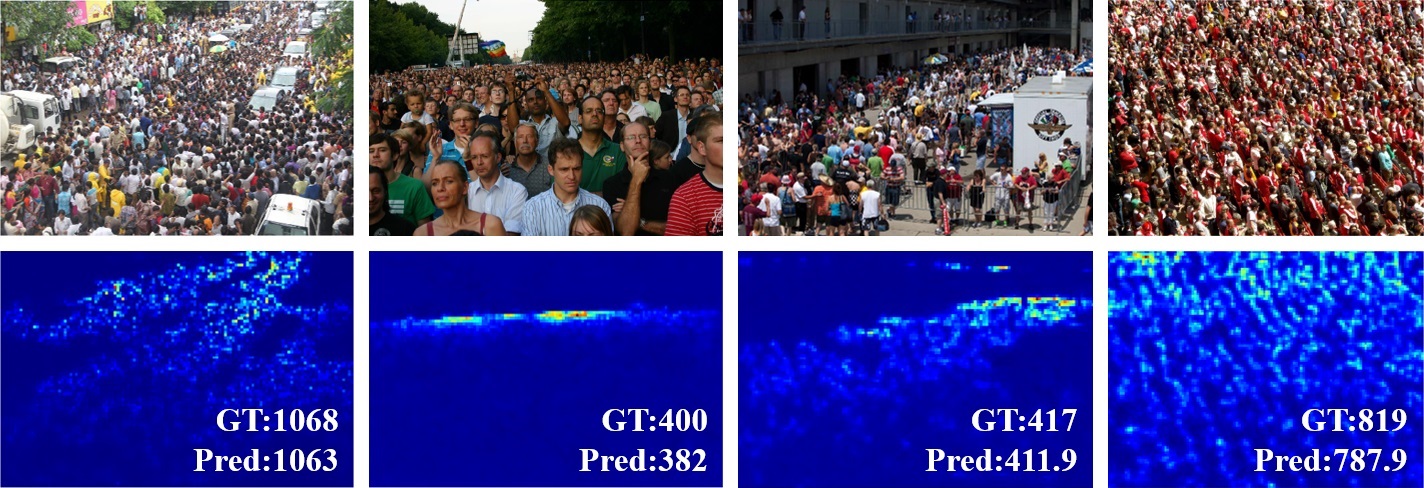}\\
  (a) ShanghaiTech Part A\\
  \vspace{0.1in}
  \includegraphics[width=.99\linewidth]{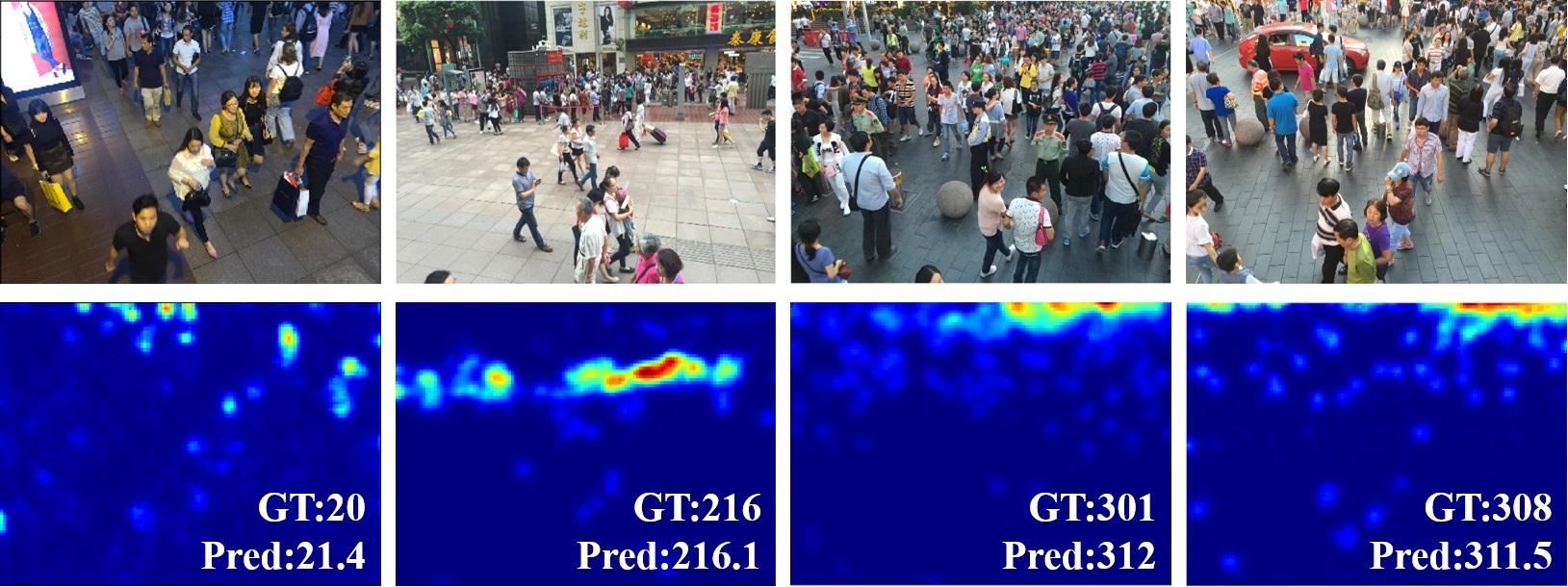}\\
  (b) ShanghaiTech Part B\\
  \vspace{0.1in}
  \includegraphics[width=.99\linewidth]{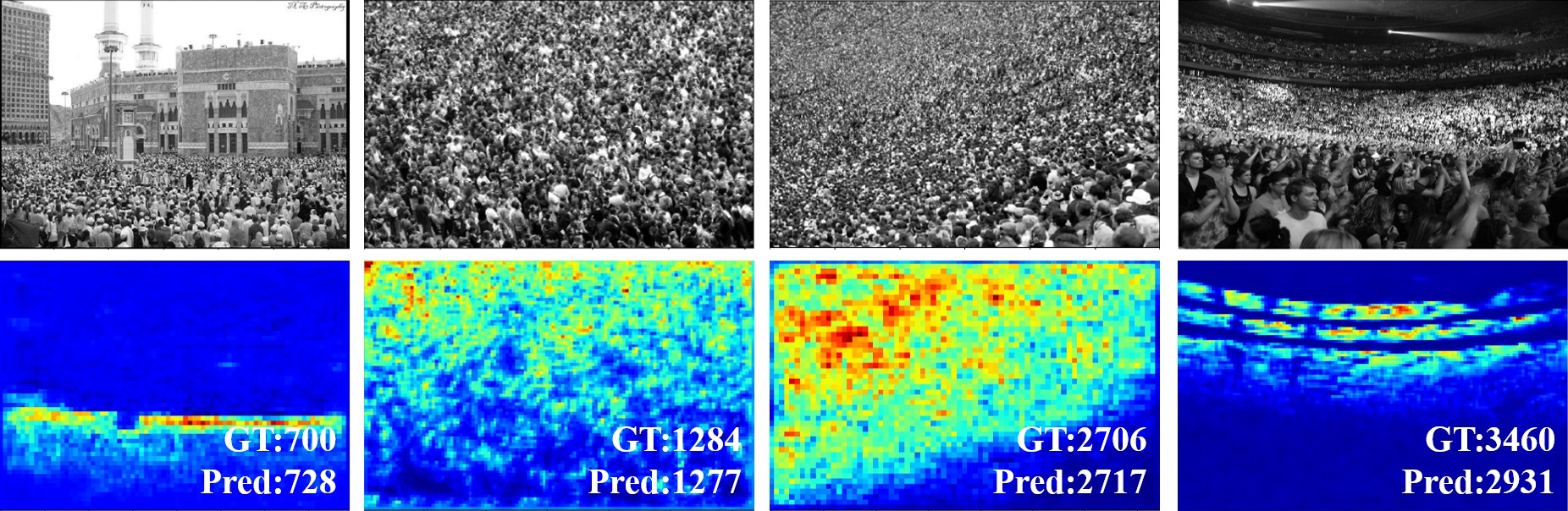}\\
  (c) UCF\_CC\_50
  \caption{Qualitative results on the benchmarks.}
  \label{fig:Qualitative}
\end{figure}

\subsection{Results and Comparison}

We first evaluate the overall results of our proposed framework. We compare our framework with several state-of-the-art approaches, including the multi-column CNN with different receptive fields~\cite{Zhang_2016_CVPR}, the Switching-CNN that leverages variation of crowd density~\cite{Sam_2017_CVPR}, and a very recent dilated convolution based model CSRNet~\cite{Li2018CSRNet}. The number of grouped scenario is 15, and the effect of the parameters will be evaluated in the next subsection. We denote our approach as \emph{ASD} (\emph{A}daptive \emph{S}cenario \emph{D}iscovery) in the following comparisons.

\vspace{0.08in}
\noindent\textbf{ShanghaiTech.}
Table~\ref{table:mae} summarizes the MAE and MSE of previous approaches and ours in the datasets.
On Part A of ShanghaiTech, we achieve a significant overall improvement of 24.8 of absolute MAE value over Switching-CNN~\cite{Sam_2017_CVPR} and 2.6 of MAE over the state-of-the-art CSRNet~\cite{Li2018CSRNet}.
On Part B, our ASD framework also achieves the best MAE 8.5 and MSE 13.7 compared to the state-of-the-art.
Fig. \ref{fig:Qualitative}(a) and (b) illustrate the density maps and the prediction results of some crowd images from both parts respectively.

\vspace{0.08in}
\noindent\textbf{UCF\_CC\_50.}
We now report results on the UCF\_CC\_50 dataset, as summarized in Table \ref{table:mae} and shown in Fig. \ref{fig:Qualitative}(c).
Similar to the experiments on ShanghaiTech, the ASD framework shows better results than the other approaches, and we improve on the previously reported state-of-the-art results by 26.3\% for the MAE metric and 31.8\% for the MSE, which indicates the low variance of our prediction across the high crowd density images.

\begin{figure}[t]
  \centering
  \includegraphics[width=0.9\linewidth]{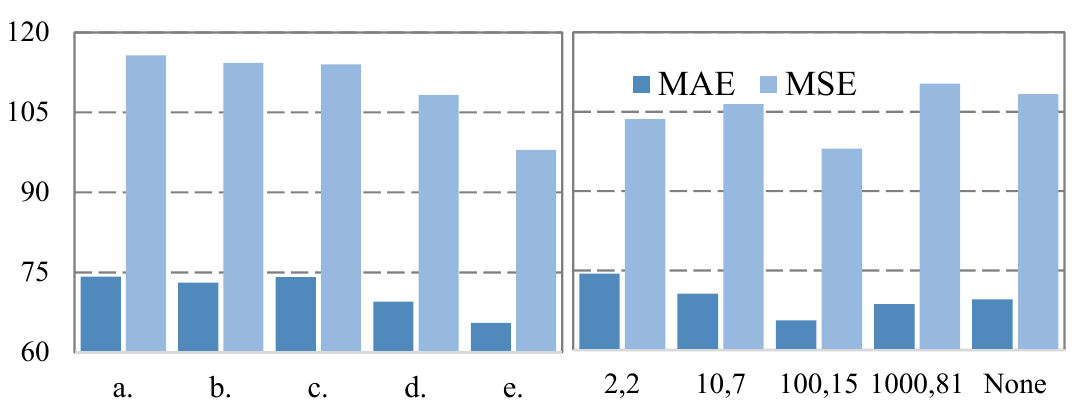}
  \caption{Left: the effect of varying network architecture, a. sparse pathway only; b. dense pathway only; c. fusion of the two pathways with the same weight; d. learned weight without discretization; e. proposed approach. Right: the effect of scenario discovery w.r.t the number of discretization bins and grouped scenarios (``None'' indicates the result without discretization).}
  \label{fig:sdcomparison}
\end{figure}

\begin{figure}[t]
  \centering
  \includegraphics[width=.95\linewidth]{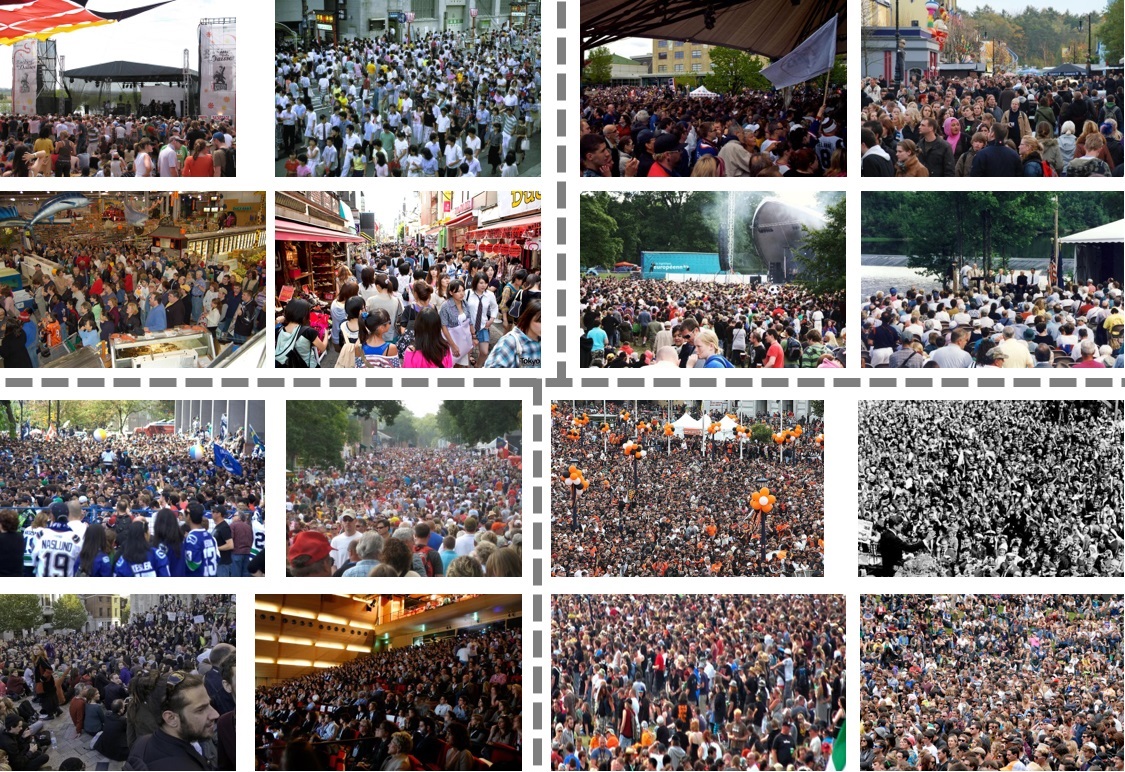}
  \caption{Images of four sample scenarios grouped by adaptive scenario discovery. A various of differences between each two scenarios, such as crowd density, ratio of background, and viewpoints, can be visibly from the images.}
  \label{fig:scenario}
\end{figure}

\subsection{Ablation Study}

In this part, we evaluate a few parameters and an alternative implementation for the proposed framework.
We report results on the ShanghaiTech Part A.

\vspace{0.08in}
\noindent\textbf{Network Architecture.}
We first evaluate the effect of the two parallel pathways over the whole framework.
Fig. \ref{fig:sdcomparison}-Left gives the comparison with different network architecture, including the single pathway and the fusion of them. With the fusion of a fixed pathway weight, the result is 74.1 of MAE and 114.0 of MSE, which is not higher than results by the single pathway. We observe significant performance gains when adding the dynamic pathway-wise responses and the discretization.

\vspace{0.08in}
\noindent\textbf{The Effect of Scenario Discovery.}
Recall that the discretization on the adaption branch is applied to discover the dynamic scenarios implicitly; here we consider the different choice of parameters. The output response after the operation of Equ. (1) fall in the interval (0,1), and is divided into 2,10,100, and 1000 bins. Note that only a proportion of bins are with images after model training due to the size of the dataset, therefore the number of scenarios is usually smaller than that of the bin. Discretization with 2 bins can be considered as a simplified version of Switching-CNN~\cite{Sam_2017_CVPR}, and our learning strategy still achieves lower MAE (74.4 vs. 90.4). Without the discretization, we obtain the MAE of 69.4, which is not as good as the scenario discovery with 15 and 81 scenarios (MAE of 65.6 and 68.7, respectively). Fig. \ref{fig:scenario} shows some crowd images from different scenarios.

\section{Conclusions}
\label{sec:conclusion}
In this paper, we have presented a novel architecture for high-density population counting. Our approach focuses on the implicit discovery and dynamic modeling of scenarios. In addition, we have reformulated the crowd counting problem as a scenario classification problem such that the semantic scenario models into a combined prediction sub-tasks. The adaptive scenario discovery is built to obtain two weights of different sizes through the parallel perception path for dynamic fusion. Our proposed framework achieves state-of-the-art performance on two popular crowd counting datasets.

{
\balance
\bibliographystyle{IEEEbib}
\bibliography{total}
}

\end{document}